# Severity and Mortality prediction models to triage Indian COVID-19 patients

**Authors**

Samarth Bhatia[1,¶], Yukti Makhija[2,¶], Sneha Jayaswal[4], Shalendra Singh[3], Ishaan Gupta[2]*

**Affiliations**
[1]Department of Chemical engineering, Indian Institute of Technology, Delhi, Hauz Khas, New Delhi-110016.
[2]Department of Biochemical Engineering and Biotechnology, Indian Institute of Technology, Delhi, Hauz Khas, New Delhi-110016.
[3]Department of Anaesthesiology and Critical Care, AFMC, Pune-411040, Maharashtra.
[4]CMC Ludhiana, Ludhiana-141008, Punjab

* Corresponding author
Email: ishaan@iitd.ac.in (IG)

[¶]These authors contributed equally to this work.

**Abstract**

As the second wave in India mitigates, COVID-19 has now infected about 29 million patients countrywide, leading to more than 350 thousand people dead. As the infections surged, the strain on the medical infrastructure in the country became apparent. While the country vaccinates its population, opening up the economy may lead to an increase in infection rates. In this scenario, it is essential to effectively utilize the limited hospital resources by an informed patient triaging system based on clinical parameters. Here, we present two interpretable machine learning models predicting the clinical outcomes, severity, and mortality, of the patients based on routine non-invasive surveillance of blood parameters from one of the largest cohorts of Indian patients at the day of admission. Patient severity and mortality prediction models achieved 86.3% and 88.06% accuracy, respectively, with an AUC-ROC of 0.91 and 0.92. We have integrated both the models in a user-friendly web app calculator, https://triage-COVID-19.herokuapp.com/, to showcase the potential deployment of such efforts at scale.

**Author Summary**

As the medical system in India struggles to cope with more than 1.5 million active cases, with a total number of patients crossing 30 million, it is essential to develop patient triage models for effective utilization of medical resources. Here, we built cross-validated machine learning models using data from one of the largest cohorts of Covid-19 patients from India to categorize patients based on the severity of infection and eventual mortality. Using routine clinical parameters measured from patient blood we were able to predict with about 90% accuracy the progression of disease in an individual at the time of admission. Our model is available as a web application https://triage-covid-19.herokuapp.com/ and is easily accessible and deployable.

1. **Introduction**

Translational science is a rapidly growing field with immediate potential in direct clinical applications. This includes the development of computational models that analyze Electronic Health Records (EHRs) and aid us in interpreting the complex biological associations between clinical measurements and patient outcomes. Machine Learning is an extremely powerful tool deployed by translational scientists to recognize patterns and identify features from medical data that correlate with clinical outcomes. The predictions obtained from such machine learning models may assist in clinical decision-making. This can enable us to automate certain stages of diagnosis, especially when during a scarcity of medical resources such as trained medical professionals or intensive care units (ICUs).

Globally, several models have been proposed by researchers to tackle the problem of triaging COVID-19 patients to budget for, allocate and effectively manage appropriate medical resources, such as by D Ellinghaus et al. (4) and by Yuan Hou et al. (5). These studies found that the susceptibility and mortality rates are widely variable across countries. However, only a handful of these models have been trained on datasets populations in the developing world, such as India. To the best of our knowledge, only two such models have been trained and tested on Indian populations. First, a model based on a Random Forests classifier predicting ICU admissions using features such as age, symptoms at diagnosis, number of comorbidities, chest X-ray, SpO2 concentration, ANC/ALC ratio, CRP, and Serum ferritin concentrations (1). Second, a set of models forecast the number of COVID-19 cases in India using simple SEIR mathematical models and also using LSTMs (2, 3).

## 2. Results

Li Yan et al. proposed one of the first mortality prediction models for COVID-19 (7). This model was trained and tested on 375 infected patients in the region of Wuhan, China. Of the 375 patients, 201 had recovered, and 174 had died. This paper also proposed a clinically operable decision tree that predicted the outcome based on lactic dehydrogenase (LDH), lymphocytes, and high-sensitivity C-reactive protein (hs-CRP) values. They achieved 100% accuracy in predicting COVID-19 severity and 81% accuracy in predicting patient mortality in their dataset using this decision tree. However, as we demonstrated previously (8), testing these models on our cohort of Indian patients was not as successful. Applying this model to a subset of 120 patients from the current cohort of patients, ensuring a maximum overlap of parameters used in the model trained by Li Yan et al., the overall severity prediction accuracy was 65.26%, and the mortality prediction accuracy was 88% (8). The poor performance in predicting severity is of particular concern as this directly affects the expected medical resources that a patient may require, thereby fulfilling the purpose of triaging. Therefore, it was noted that the existing models might be population-specific due to various intrinsic factors such as genetics and external factors such as demography, population density, or access to appropriate medical infrastructure.

Following this, we endeavored to build two supervised machine learning models pertaining to the needs of the Indian population.

## 2.1 Evaluation of the Mortality Prediction Model

| Training Dataset | 302 patients (165 alive + 137 dead) |
|---|---|
| Validation Dataset | 73 patients (40 alive + 33 dead) |
| Training Accuracy | 92.72% |
| Validation Accuracy | 86.30% |
| F-score | 0.8485 |
| Sensitivity (Recall) | 0.8485 |
| Specificity | 0.875 |
| PPV (Precision) | 0.8485 |
| NPV | 0.875 |
| AUC-ROC (on validation set) | 0.91 |
| AUC-PRC (on validation set) | 0.88 |

**Table 1:** Performance of the mortality prediction model

From the SHAP(13) value and the mean |SHAP| value importance plot (*Figure 1*), we can get an idea about the relative importance of different parameters. We see that age plays a vital role in determining the mortality of a patient, which is in accordance with the literature available on COVID-19. Having a lower value for age greatly impacts the model's output negatively (or favors the negative class, i.e., alive) than the positive class (which is deceased in this case). According to the Union Health Ministry, Government of India, ~53% of people who have died because of COVID-19 in India are above the age of 60 (6).

Other parameters that are useful to determine the mortality of a patient are the percentage of neutrophils in the blood, the creatinine levels in the urine, and the Urea levels as well, all of which, when present in higher amounts than usual, point towards a positive classification. Parameters moderately impacting the output are alkaline phosphatase (ALP) enzyme level, serum sodium levels in the blood, and indirect bilirubin in the blood, suggesting that the liver could be affected by COVID-19. We can see from the ROC curve, plotted on the validation set, that it covers an area of 0.91, which means it can separate the classes correctly 91% of the time (***Figure 1***).

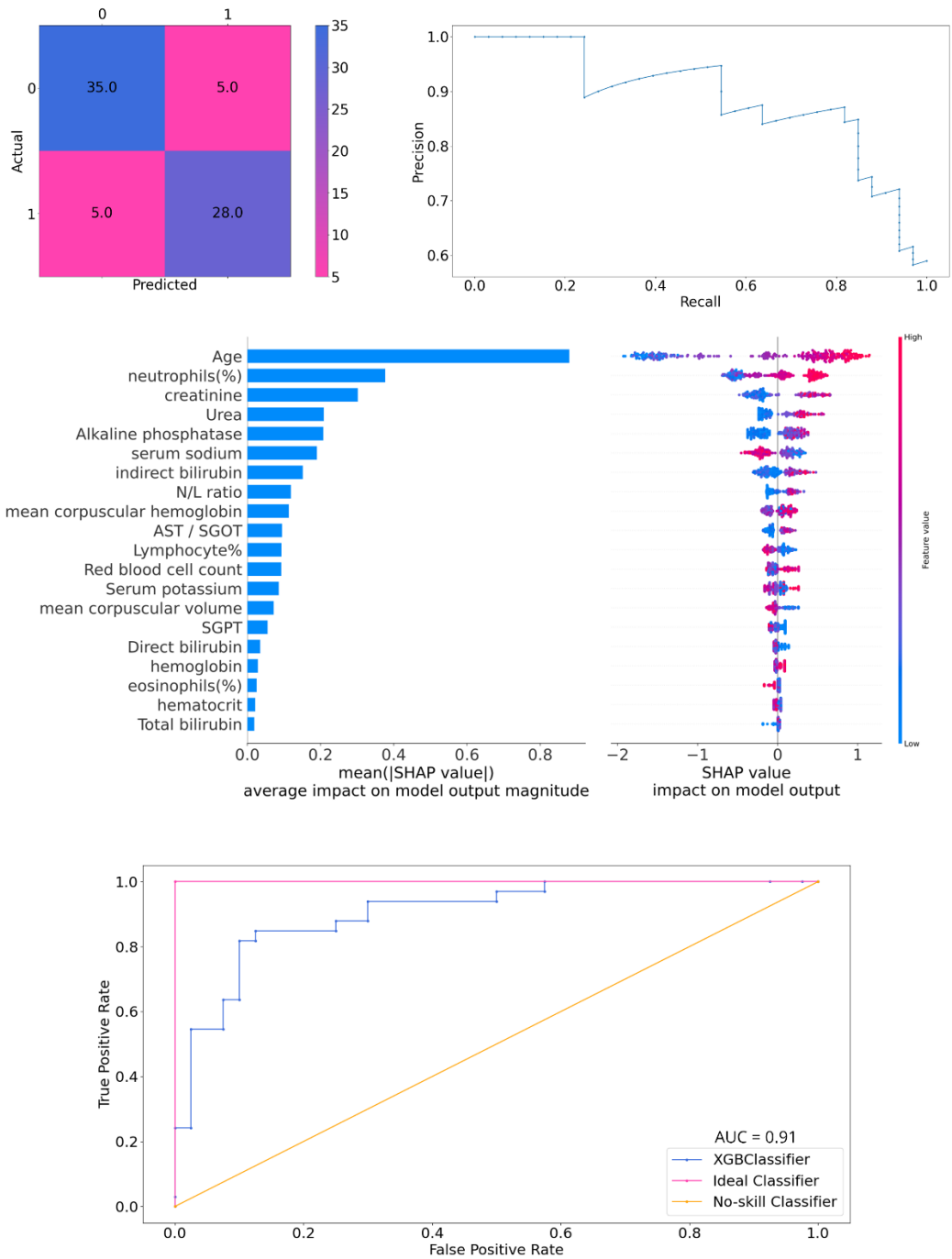

**Figure 1:** (**A**) Confusion matrix for the mortality model (**B**) Precision-Recall curve for the mortality model (**C**) Description of SHAP values for parameters with average impact on the model (left) and distribution of feature values (right) (**D**) ROC curve for the model

## 2.2 Evaluation of the Severity model

| Training Dataset | 264 patients (146 severe + 118 non-severe) |
|---|---|
| Validation Dataset | 67 patients (37 severe + 30 non-severe) |
| Training accuracy | 91.67% |
| Validation accuracy | 88.06% |
| F-score | 0.892 |
| Sensitivity (Recall) | 0.892 |
| Specificity | 0.867 |
| PPV (Precision) | 0.892 |
| NPV | 0.867 |
| AUC-ROC (on validation set) | 0.92 |
| AUC-PRC (on validation set) | 0.93 |

**Table 2:** Performance of severity prediction model

We see from the plots of the SHAP and the mean |SHAP| values (Figure 2), Age and Urea levels play an essential role in determining the severity of the disease. But, since we had used additional biomarkers for COVID-19 in this model, we get interesting results showing that High sensitivity C-reactive protein (hs-CRP) and D-D dimer have a big impact on the model. We see that for all these parameters, an increased value suggests that the patient is categorized as severe. However, we also see indirect bilirubin and AST/SGOT having a moderate impact on the model output as well. The severity model performs well with its ROC curve (plotted on validation set) covering an area of 0.92, thus differentiating between classes 92% of the time, slightly better than the mortality prediction model. A comparison of the two models is made in the next section.

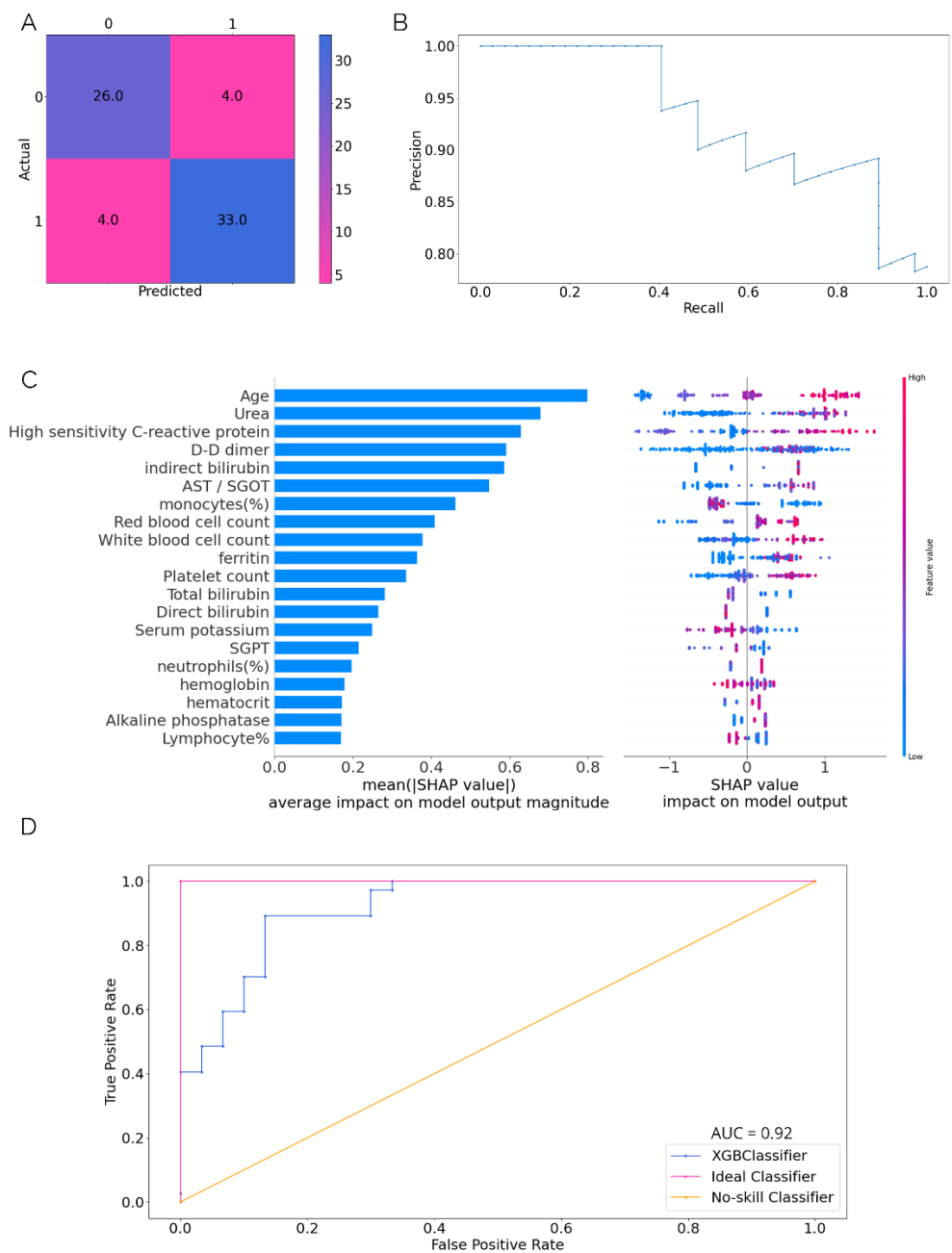

**Figure 2:** (**A**) Confusion matrix for the severity model (**B**) Precision-Recall curve for the severity prediction model (**C**) Description of SHAP values for parameters with average impact on the model (left) and distribution of feature values (right) (**D**) ROC curve for the model

**2.3 Reduced Models**

Since our machine learning models use 33 clinical parameters, which is a lot and could cause problems because of data unavailability. The distributions of the important features (according to the feature importance plots) are given in Supplementary Material (Section 3: Supplementary Figure 2). We decided to make 'reduced' models that use only the top ten features that were found by the mean SHAP values since ten features would be easier and faster to collect from a patient than 33 features. We performed hyperparameter tuning again for the two reduced models of mortality and severity and were able to achieve good accuracy and similar AUC-ROC score of 0.91 and 0.93, respectively, on the validation set without facing overfitting (***Supplementary Section 4: Supplementary Table 3,4***).

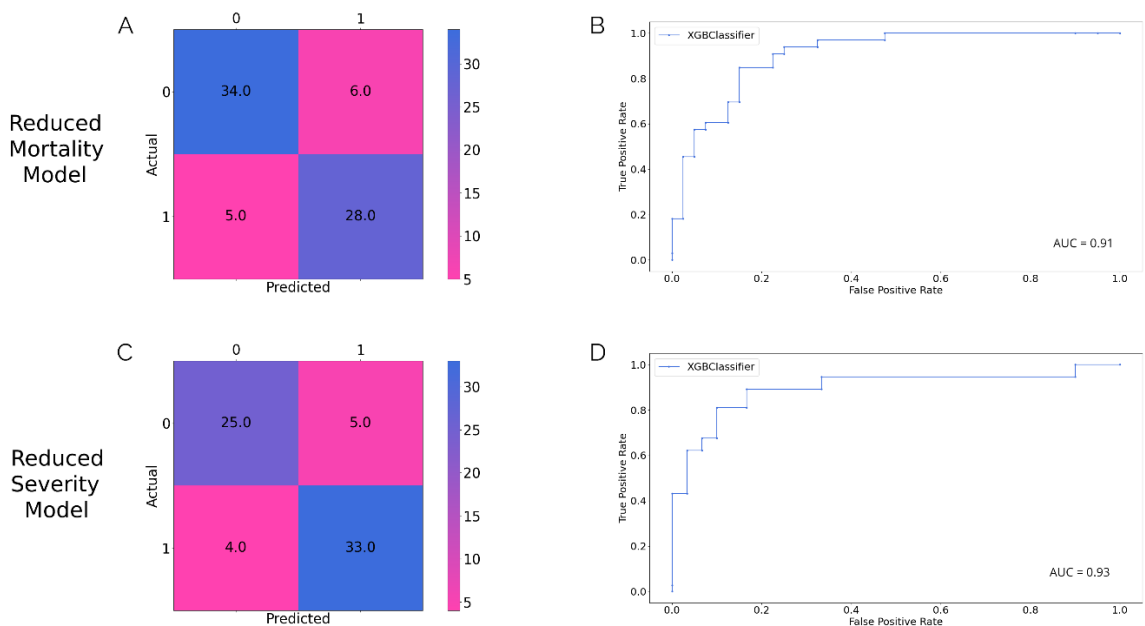

**Figure 3:** Confusion matrixes and ROC curve for the Reduced Mortality (top) and Severity prediction (bottom) models

## 3. Discussion and Conclusion

We were successfully able to make two machine learning models to triage COVID-19 patients in India into prediction categories like alive versus deceased and severe versus non-severe. This was done using XGBoost (12) models that use gradient boosting in decision trees (the `gbtree` booster of XGBoost was internally used). These models can be summarized through representative decision trees (Figure 4) to inform a clinical decision support system. In order to make the results of our model available for testing, we made an online calculator web app that can be used to determine the mortality and severity using our models at: https://triage-COVID-19.herokuapp.com/.

In the mortality model, we see that most of the evaluation parameters, like test accuracy, F-score, recall, precision, and AUC-ROC scores are less than those for the severity model. This can be because, for the severity model, we have used certain biomarkers that have been shown to be affected by COVID-19. These include high sensitivity C-Reactive Protein (hs-CRP), ferritin, D-D dimer, and Lactate Dehydrogenase (LDH). However, we could not use LDH because of data limitations as described in the section on building the severity model. Even so, we found in the above section that ferritin does not contribute as much to the severity model's output as hs-CRP and D-D dimer do. This could imply there is a more direct correlation between COVID-19 and hs-CRP and D-D dimer than ferritin. Additionally, in the ROC curves, the severity model can achieve around 60% sensitivity (TPR) with a very low FPR, whereas the mortality model can only achieve up to about 50% sensitivity for the same low FPR. In the precision-recall curves, we see that the mortality model can achieve a recall of 0.2 without predicting any false positives, whereas severity can achieve a recall of 0.4.

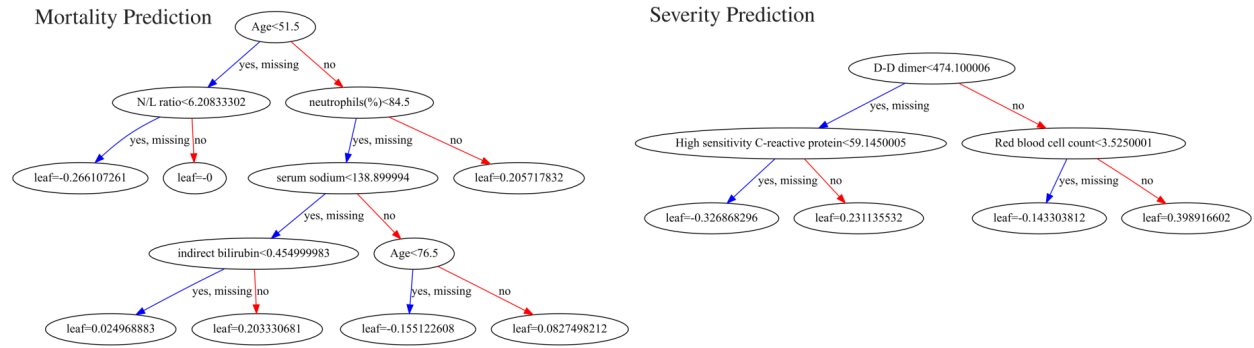

**Figure 4:** Representative decision trees of mortality (left) and severity (right) prediction models to test clinical applicability

## 4. Materials and Methods

### 4.1 Data Collection

The data for this study was collected from one of the largest dedicated COVID-19 centers in Northern India, Sardar Vallabhbhai Patel COVID Hospital and PM CARES COVID Care Hospitals. In these hospitals, a series of blood tests were performed for all the patients with confirmed infection at the time of admission. The patients were diagnosed using a Rapid Antigen Test or RT-PCR testing of a nasal/throat swab sample. The data for 815 patients were collected between 13th July and 31st December 2020. The clinicians followed the guidelines provided by the MOHFW, Govt. of India(9), to classify 390, 160, 84, 181 patients into the categories mild, moderate, severe, and dead based on symptoms on arrival. These results were used to identify the primary biomarkers that control the risk of developing severe infections and survival chances.

### 4.2 Data Preprocessing

Data preprocessing involved removing missing values by deleting the patients or features entirely if they are available for significantly fewer patients. We avoided simple imputation, i.e., replacing missing values by the mean, because the size of the dataset was not appreciable enough to guess values accurately. Imputing the missing values would have resulted in a low variance, and decreasing our data's variance directly implies adding a bias to our model. Other than this, we converted all categorical variables into one-hot encoded form, and we made binary variables for signs/symptoms and comorbidities of each patient. The comorbidities were classified into broad categories: cardiac disease, chronic liver disease, hypertension, diabetes, chronic kidney disease, lung disease, morbid obesity, and hypothyroidism. The dataset obtained after completing data-preprocessing contained 600 patients, out of which 170 were deceased, 250 had experienced mild symptoms, 99 moderate, and 81 severe with 33 clinical parameters.

**4.3 Clustering**

We had performed clustering of patients based on their collected clinical parameters using the KPrototypes algorithm to check for the presence of any clinical bias. We have summarized the results in Supplementary Material (Section 1).

**4.4 Modelling Strategy**

**4.4.1 Mortality Prediction Model**

All the features, including their gross statistics, used for training mortality prediction models can be found in Supplementary Material (Section 2: Supplementary Table 1). The number of deceased patients was less than those who had recovered (170:430); this formed a skewed data distribution. Standard machine learning algorithms perform poorly on such datasets as they tend to be biased towards the majority class. This decreases the prediction accuracy of the minority class (10). For this reason, we decided to use ensemble models, which are multiple classifier methods, i.e., these learning machines combine the decision of multiple base classifiers to reach the final prediction. A widely-used class of ensemble models is boosting algorithms, in which the final classifier is built through the sequential addition of multiple weak classifiers. The overall performance of the model increases with the addition of each classifier. We did an extensive survey of the different CART (Classification and Regression Trees) models and decided to implement AdaBoost(16), XGBoost(12), and CatBoost(15) because of their proof of performance. In general, we found that the validation AUC-ROC of XGBoost (0.91) was significantly more than CatBoost (0.80) and AdaBoost (0.72).

In boosting algorithms, a 1:1 ratio of the classes in the data is recognized as ideal. We used random undersampling for the majority class, i.e., recovered patients, to reduce the data imbalance. The new dataset contained the information of 170 deceased patients and 205 recovered patients. We ensured that the number of mild, moderate, and severe patients were equally represented (Mild-55, Moderate-55, Severe-55) in the dataset to minimize bias in the model. Following this, we split this dataset into training and validation sets containing 302 and 73 patients, respectively. The splitting was performed using stratified sampling to maintain the ratio of mild, moderate, severe, and deceased patients in both sets.

Once the training dataset had been finalized, we started with the training of the model. We implement Repeated Stratified 5-fold cross-validation using the GridSearchCV function from the sklearn(14) python package. While iterating with a wide range of hyperparameters during cross-validation, we have optimized the model for the best accuracy and re-fitted it for the best F-Score. The hyperparameter values of the model obtained after cross-validation are listed in **Table 3** and the cross-validation results are listed in **Table 4**. Using colsample_bytree enables us to induce randomness in the training examples and make our model robust from noise, and varying max_depth, min_child_weight, gamma, and alpha enables us to reduce and remove any overfitting.

### 4.4.2 Severity Prediction Model

All the features, including their gross statistics, used for training mortality prediction models can be in Supplementary Material (Section 2: Supplementary Table 2). The objective of this model was to predict the risk of developing severe infection. We clubbed mild and moderate patients as non-severe, and severe and dead patients were grouped as severe. Many researchers have found that COVID-19 causes a change in these four critical biomarkers: D-D Dimer, High Sensitivity C-Reactive Protein (hs C-RP), Lactate Dehydrogenase (LDH), and ferritin. However, these were not measured for all patients. We had 396, 390, 305, 398 patients for those parameters, respectively. We calculated the number of unique patients with different combinations of three of these parameters and found the maximum to be 331 patients with values for D-Dimer, hs-CRP, and ferritin as the optimal combination keeping in mind how the model performance depends on the quantity of data and number of parameters. Following this, we split these 331 patients into training and validation sets containing 264 and 67 patients, respectively. The choice of model was the same as in the mortality prediction model mentioned above. After hyperparameter tuning and cross-validation as mentioned in the mortality model above, the hyperparameters of the best model are mentioned in **Table 3**. The results are cross-validation are listed in **Table 4**.

### 4.4.3 Evaluation Methodology

We evaluated the models based on different parameters, namely Accuracy, Recall, Precision, and F-Score, along with the ROC and Precision-Recall curves. These are defined in terms of the number of True Positives (TP), False Positive (FP), True Negative (TN), and False Negative (FN) for a class as:

$$\text{Accuracy} = \frac{TP + TN}{TP + TN + FP + FN}$$

$$\text{Precision} = \frac{TP}{TP + FP}$$

$$\text{Recall/True Positive Rate (TPR)/Sensitivity} = \frac{TP}{TP + FN}$$

$$\text{F-Score} = \frac{2 \times Precision \times Recall}{Precision + Recall}$$

**Specificity/True Negative Rate (TNR)** = $\frac{TN}{TN+FP}$

The **ROC (Receiver Operating Characteristics)** curve can be made by plotting **TPR** (Recall) vs. **FPR** (1 – specificity (or TNR)). It is a measure of how well our model can differentiate between 2 classes. The **AUC-ROC** is the Area under the ROC curve.

The methods `accuracy_score()`, `recall_score()`, `precision_score()` and `f1_score()` of the sklearn.metrics module were used to evaluate the respective parameters and the `plot_roc_curve()` and `precision_recall_curve()` to plot the ROC and P-R curves and get the AUC value.

For the mortality model, the positive class was chosen to be deceased(dead) patients and the negative class to be alive (mild + moderate + severe) patients. Following the same pattern, in the severity model, the positive class was chosen to be severe (severe + dead) patients, whereas the negative class was chosen to be non-severe (mild + moderate) patients.

The SHAP value plots are not part of the model itself; rather, SHAP uses an external model to vary the value of every feature and observe how much the output is affected. The SHAP value plots tell us how these variations affect the model's output. The mean |SHAP| value plot gives us information about which features are the most important in our model and have the most impact.

Please note that we can use the precision-recall curve here since we are more concerned with the positive class and having fewer False Negatives than False Positives (as patients whose mortality/severity goes undetected on being classified as non-severe/alive is more harmful to the society than a patient which was not severe being classified as severe/dead) (11).

| Hyperparameters | Mortality Prediction Model | Severity Prediction Model |
|---|---|---|
| alpha | 0.9 | 0.1 |
| gamma | 0.8 | 3 |
| n_estimators | 100 | 100 |
| min_child_weight | 2 | 1 |
| subsample | 1 | 0.3 |
| colsample_bytree | 0.7 | 1 |
| learning_rate | 0.148 | 0.3 |
| max_depth | 4 | 6 |

**Table 3:** List of Hyperparameters used in the Models

| Model | Mean Accuracy (95% CI) | Mean AUC-ROC (95% CI) |
|---|---|---|
| **Mortality** | 79.74% (70.33%, 89.15%) | 0.883 (0.807, 0.959) |
| **Severity** | 80.31% (75.32%, 85.29%)) | 0.888 (0.812, 0.964) |

**Table 4:** Results of Cross-Validation (5-Fold, Stratified, 100 Iterations)

**Confidentiality statement and ethics**

The study protocol was reviewed and approved by Institute ethics committees at both the participating centres through Institute ethics committee File No. IEC/320/325 to SS. Consent waiver was granted given the retrospective nature of analysis and emergent nature of the pandemic. Confidentiality was maintained by the de-identification of data. All analysis was performed on de-identified data.

**Acknowledgment**

We would like to acknowledge the nursing staff and medical professionals who have tirelessly worked to alleviate the suffering caused by this pandemic and facilitated the collection of good-quality data for this publication.

**Financial support & sponsorship**

This research did not receive any specific grant from funding agencies in the public, commercial, or not-for-profit sectors.

**Conflicts of Interest**

The authors have declared that no competing interests exist.

**Data and Code Availability**

We have uploaded the de-identified data to a GitHub repository: https://github.com/yuktimakhija/COVID-19-Patient-Triaging

# 6. Supplementary Material

## 6.1 Clustering to visualize clinical bias

For maintaining uniformity and removing skewness from the distributions of the features, we transformed the data using `PowerTransformer` of the `sklearn.preprocessing` module, which implements the Yeo-Johnson transformation. (1)

The clustering was done using the KPrototypes(2) algorithm implemented in the `kmodes` package using the number of clusters as 2 since we had two classes (severe and non-severe). The `umap` (3) and `plotly.express` (4) packages were used to reduce the dimensionality of the data and make the plots, respectively.

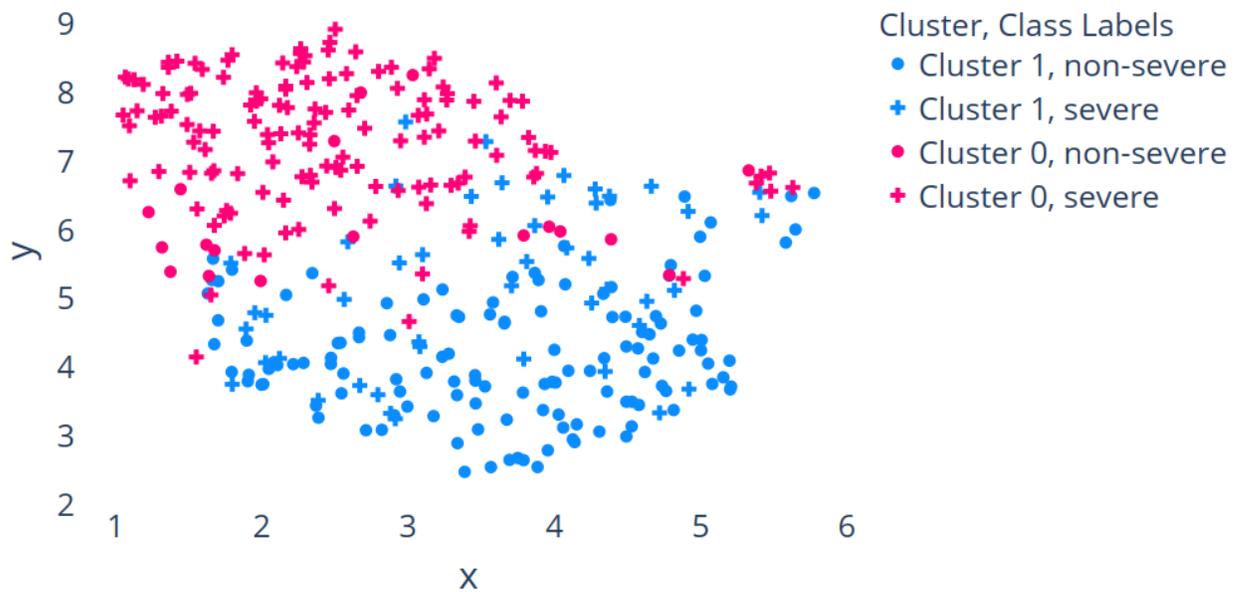

**Supplementary Figure 1:** The points are coloured according to the clustering labels (learned from data), and the point shape represents the severity status of the patients.

## 6.2 Description of gross clinical parameters collected.

| Features | Alive Patients | Deceased Patients |
| --- | --- | --- |
| | Median (Interquartile Range) | Median (Interquartile Range) |
| AST / SGOT (U/L) | 41.4 (36.76, 65.1) | 55.1 (36.76, 86.10) |
| Age (Years) | 49 (36, 60) | 70 (62, 79) |
| Alkaline phosphatase (U/L) | 92.4 (75.15, 119.8) | 109.77 (83.53, 134.93) |
| Direct Bilirubin (mg/dL) | 0.28 (0.2, 0.39) | 0.3 (0.21, 0.41) |
| Lymphocyte (%) | 21 (13, 29) | 9 (5, 15) |
| N/L Ratio | 3.3 (2.14, 6.08) | 9.44 (5.06, 17.9) |
| Platelet Count ($10^5$ cells/µL) | 1.62 (1.16, 2.38) | 1.85 (1.35, 2.77) |
| RBC Count (cells/mcL) | 4.41 (4, 4.82) | 4.42 (3.98, 4.85) |
| SGPT (U/L) | 39.6 (24.5, 70) | 41.38 (21.95, 66.07) |
| Serum Potassium (mmol/L) | 4.5 (4.03, 4.91) | 4.5 (4.03, 5.05) |
| Total Bilirubin (mg/dL) | 0.65 (0.49, 0.84) | 0.78 (0.54, 1.03) |
| Urea (mg/dL) | 25.1 (19.8, 32.6) | 48.05 (31.18, 80.88) |
| WBC Count (cells/µL) | 6800 (5300, 8900) | 11300 (7200, 17950) |
| Basophil (%) | 0 (0, 0) | 0 (0, 0) |
| Creatinine (mg/dL) | 1 (0.87, 1.16) | 1.33 (1.0, 1.68) |
| Eosinophils (%) | 2 (1, 3) | 1 (1, 2) |
| Hematocrit (%) | 39.5 (35.5, 42.5) | 38.1 (34, 42.3) |
| Hemoglobin (g/dL) | 12.5 (11.3, 13.8) | 12.25 (10.9, 13.98) |
| Indirect Bilirubin (mg/dL) | 0.35 (0.24, 0.48) | 0.42 (0.28, 0.67) |
| Mean Corpuscular Hemoglobin (pg) | 28.7 (26.9, 30.7) | 28.55 (25.73, 30.98) |
| Mean Corpuscular Volume (fL) | 90.4 (85.8, 96.2) | 88.25 (81.95, 93.33) |
| Monocytes (%) | 6 (4, 10) | 4 (2, 6) |
| Neutrophils (%) | 69.9 (60, 79) | 86 (77.25, 90) |
| Outcome (0: Deceased 1: Alive) | 0 (0, 0) | 1 (1, 1) |
| Serum Sodium (mmol/L) | 140 (136.1, 143.2) | 137.75 (133.85, 141.73) |
| Gender (0: M 1: F) | 0 (0, 1) | 0 (0, 1) |

**Supplementary Table 1:** Comparison of features in deceased and alive patients in the mortality prediction model

| Features (Units) | Severe Patients | Non-Severe Patients |
| --- | --- | --- |
| | Median (Interquartile Range) | Median (Interquartile Range) |
| AST / SGOT (U/L) | 37.1 (26.65, 53.98) | 51.4 (36.53, 78.08) |
| Age (Years) | 50 (41, 63.25) | 65 (55, 75) |
| Alkaline phosphatase (U/L) | 90.64 (71.35, 117.4) | 101.2 (78.48, 126) |
| Direct Bilirubin (mg/dL) | 0.26 (0.2, 0.38) | 0.3 (0.21, 0.39) |
| Lymphocyte (%) | 23 (15, 31) | 12 (7, 17.95) |
| N/L Ratio | 2.89 (1.87, 4.89) | 7 (4, 12.86) |
| Platelet Count ($10^5$ cells/μL) | 1.52 (1.06, 1.99) | 2.06 (1.44, 3.15) |
| RBC Count (cells/mcL) | 4.41 (3.96, 4.81) | 4.33 (3.94, 4.8) |
| SGPT (U/L) | 37.65 (21.2, 61.53) | 43.67 (24.21, 75) |
| Serum Potassium (mmol/L) | 4.47 (4.12, 4.9) | 4.4 (3.99, 4.89) |
| Total Bilirubin (mg/dL) | 0.58 (0.44, 0.79) | 0.68 (0.52, 0.94) |
| Urea (mg/dL) | 25.45 (21.18, 31.08) | 40.9 (28.86, 59.87) |
| WBC Count (cells/μL) | 6200 (5100, 7725) | 10900 (7010, 15750) |
| Basophil (%) | 0 (0, 0) | 0 (0, 0) |
| Creatinine (mg/dL) | 1.045 (0.91, 1.26) | 1.17 (0.95, 1.42) |
| Eosinophils (%) | 2 (1, 3) | 1 (1, 2) |
| Hematocrit (%) | 38.7 (34.98, 42.53) | 38.5 (34.25, 41.8) |
| Hemoglobin (g/dL) | 12.4 (11.2, 13.5) | 12.1 (10.85, 13.5) |
| Indirect Bilirubin (mg/dL) | 0.305 (0.21, 0.41) | 0.38 (0.27, 0.53) |
| Mean Corpuscular Hemoglobin (pg) | 28.5 (26.7, 30.5) | 28.1 (26.05, 30.25) |
| Mean Corpuscular Volume (fL) | 88.7 (84.65, 93.73) | 88.9 (82.55, 93.65) |
| Monocytes (%) | 7 (4, 10) | 3 (2, 6) |
| Neutrophils (%) | 66 (58, 76) | 83 (73.35, 89) |
| Outcome (0: Deceased 1: Alive) | 0 (0, 0) | 1 (1, 1) |
| Serum Sodium (mmol/L) | 140.65 (137.2, 143.6) | 137.5 (133.5, 141.3) |
| Gender (0: M 1: F) | 0 (0, 1) | 0 (0, 1) |
| D-D dimer (ng/mL) | 354 (241.5, 454.3) | 780 (483, 1763.55) |
| Ferritin (ng/mL) | 244 (109.93, 480.68) | 544 (260.6, 1196.25) |
| High sensitivity C-reactive protein (mg/L) | 24.75 (6.21, 47.8) | 62 (25.98, 94.63) |

**Supplementary Table 2:** Comparison of features in severe and non-severe patients in the severity prediction model

## 6.3 Description of clinical parameters used in reduced mortality prediction models

This is the supplementary document containing information about the distributions of the features that were used and were found to be of importance in the mortality and severity models. These features are:

1. D-D Dimer
2. Ferritin
3. High Sensitivity C - Reactive Protein
4. Age
5. Urea
6. Alkaline Phosphatase
7. Creatinine
8. Indirect Bilirubin
9. Neutrophil

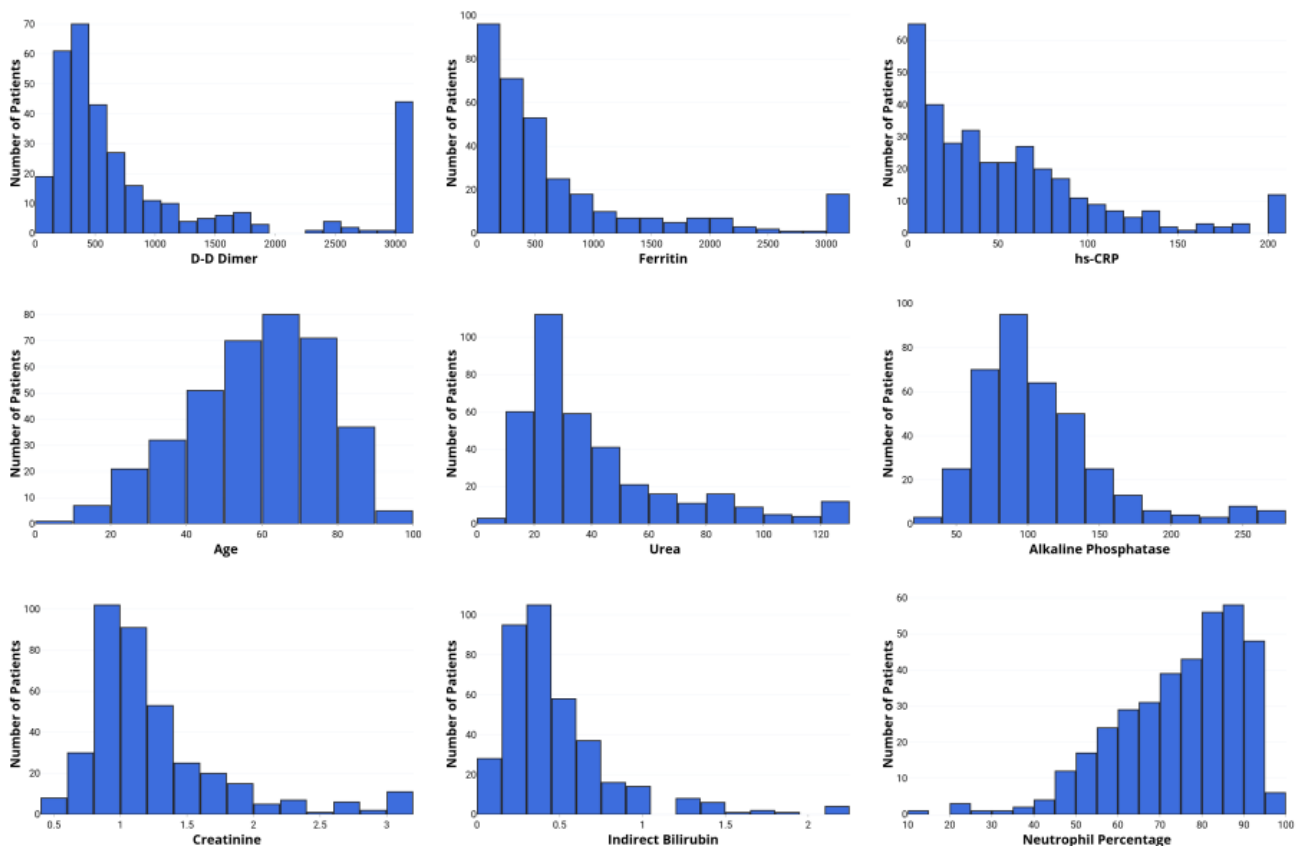

**Supplementary Figure 2:** Distribution of features with high predictive power.

## 6.4 Reduced models

### 6.4.1 Reduced Mortality prediction model:

Features used (in order of decreasing SHAP feature importance): Age, neutrophils(%), creatinine, Urea, Alkaline phosphatase, serum sodium, indirect bilirubin, N/L ratio, Mean Corpuscular Hemoglobin, and AST/SGOT

| Training Dataset | 302 patients (165 alive + 137 dead) |
|---|---|
| Validation Dataset | 73 patients (40 alive + 33 dead) |
| Training accuracy | 88.74% |
| Validation accuracy | 84.93% |
| F-score | 0.83 |
| Sensitivity (Recall) | 0.85 |
| Specificity | 0.85 |
| PPV (Precision) | 0.82 |
| NPV | 0.85 |
| AUC-ROC (on validation set) | 0.91 |

**Supplementary Table 3:** Evaluation of reduced mortality model

### 6.4.2 Reduced Severity prediction model:

Features used (in decreasing order of SHAP feature importance):
Age, Urea, High sensitivity C-reactive protein, D-D dimer, indirect bilirubin, AST / SGOT, monocytes(%), Red blood cell count, White blood cell count, ferritin

By reducing the number of features taken by the severity model and performing hyperparameter-tuning on the new model, we were able to achieve the following metrics:

| | |
|---|---|
| Training Dataset | 264 patients (146 severe + 118 non-severe) |
| Validation Dataset | 67 patients (37 severe + 30 non-severe) |
| Training accuracy | 87.12% |
| Validation accuracy | 86.57% |
| F-score | 0.88 |
| Sensitivity (Recall) | 0.89 |
| Specificity | 0.83 |
| PPV (Precision) | 0.87 |
| NPV | 0.83 |
| AUC-ROC (on validation set) | 0.93 |

**Supplementary Table 4:** Evaluation of reduced severity model

**References for Supplementary material**